# Fast Incremental Learning for Off-Road Robot Navigation

___________________________________________

**Artem Provodin**
New York University
New York NY 10003

**Liila Torabi**
Net-Scale Technologies
Morganville NJ 07751

**Beat Flepp**
NVIDIA Corp
Holmdel NJ 07733

**Yann LeCun**
New York University
New York NY10003

**Michael Sergio**
Net-Scale Technologies
Morganville NJ 07751

**L. D. Jackel**
North C Technologies
Holmdel NJ 07733

**Urs Muller**
NVIDIA Corp
Holmdel NJ 07733

**Jure Žbontar**
New York University
New York NY 10003

## Abstract

A promising approach to autonomous driving is machine learning. In such systems, training datasets are created that capture the sensory input to a vehicle as well as the desired response. A disadvantage of using a learned navigation system is that the learning process itself may require a huge number of training examples and a large amount of computing. To avoid the need to collect a large training set of driving examples, we describe a system that takes advantage of the huge number of training examples provided by ImageNet, but is able to adapt quickly using a small training set for the specific driving environment.

## 1 Introduction

Despite remarkable advances in autonomous vehicles, off-road autonomous ground navigation remains an unsolved challenge. The difficulty of the task arises from the enormous variability that an Unmanned Ground Vehicle confronts when it leaves the relatively well characterized domain of the road. While it may be possible to create a rule-based system that codifies all situations that might be presented to the vehicle, such approaches are brittle, and tend to fail when new environments are encountered.

A more promising approach to autonomous driving is machine learning. In such systems, training datasets are created that capture the sensory input to a vehicle as well as the desired response. These responses may be provided by a human driver who initially teleoperates the vehicle, or may be gleaned from the vehicle's prior driving experiences.

A number of systems have been described that learn drivable vs non-drivable terrain using handcrafted features based on color or texture (see, for example, Sofman 2010). In recent years, Convolution Neural Networks [ConvNets] (LeCun et al 1989), in which the features are learned rather than handcrafted, have been shown to be the most accurate in numerous image recognition tasks (see, for example, Krizhevsky et al 2012).

A disadvantage of using a learned system is that the learning process itself may require a huge number of training examples and a large amount of computing. As an illustration, the most accurate networks used in the ImageNet competitions (ILSVRC 2014) train on over one million examples and often require days of training on powerful processor clusters. While such expensive training may be fine for some applications, it is unacceptable for systems that have to adapt quickly and only have limited training data. Here we describe a system that takes advantage of the huge number of training examples provided by ImageNet, but is able to adapt quickly using a small training set.

**2 System Overview**

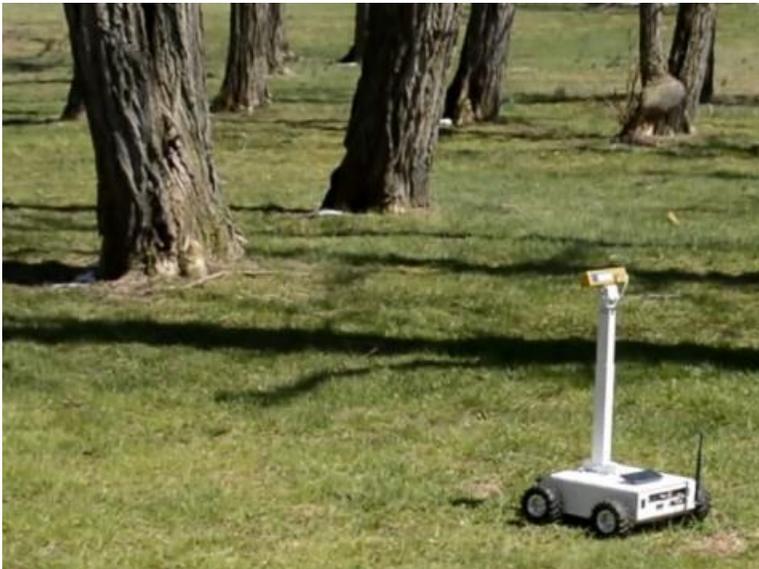

Figure 1. The CoroBot robot in a test environment: Bayonet Farm, Holmdel NJ.

Our experiments were performed using a CoroBot Jr robot as a test platform. The Corobot is skid-steered by independently varying the speeds of the four wheel motors. Our Corobot was augmented with a Point Grey Bumblebee stereo camera, a GPS (on a cell phone), and an NVIDIA Jetson GPU/CPU board. The camera was mounted on an extension stalk to provide a better view of the terrain. See Figure 1. The various components communicated with each other using the framework of the Robot Operating System (ROS).

Our CoroBot was capable of traversing obstacles that were a few centimeters high, but it could get stuck in thick or wet grass.

The navigation system consisted of a terrain classifier, a cost map, and a simple path planner. A motor control unit used the output of the path planner to issue driving commands to the CoroBot motors.

The primary emphasis of our research was in developing a terrain classifier that could rapidly adapt to new environments and thus our discussion in this paper focusses on the classifier.

To detect the initial set of obstacles, images from the stereo camera were processed to create a 3D point cloud. A ground plane was then calculated using the 3D Hough transform plane fitter. After the ground plane was computed all elements from the point cloud were labeled as either obstacle or traversable based on their distance to the ground plane. From the labeled point cloud, a cost map and labeled images are created.

The navigation system calculates the optimal driving direction based on the cost map and the goal position. The trajectory planner incorporates the distance to an obstacle to calculate the drive direction so that the CoroBot avoids obstacles.

Figure 2 shows an example of terrain labeling from stereo. The left image shows the terrain labeled using stereo and the right image shows the resultant cost map. The map is about 15 meters by 15 meters.

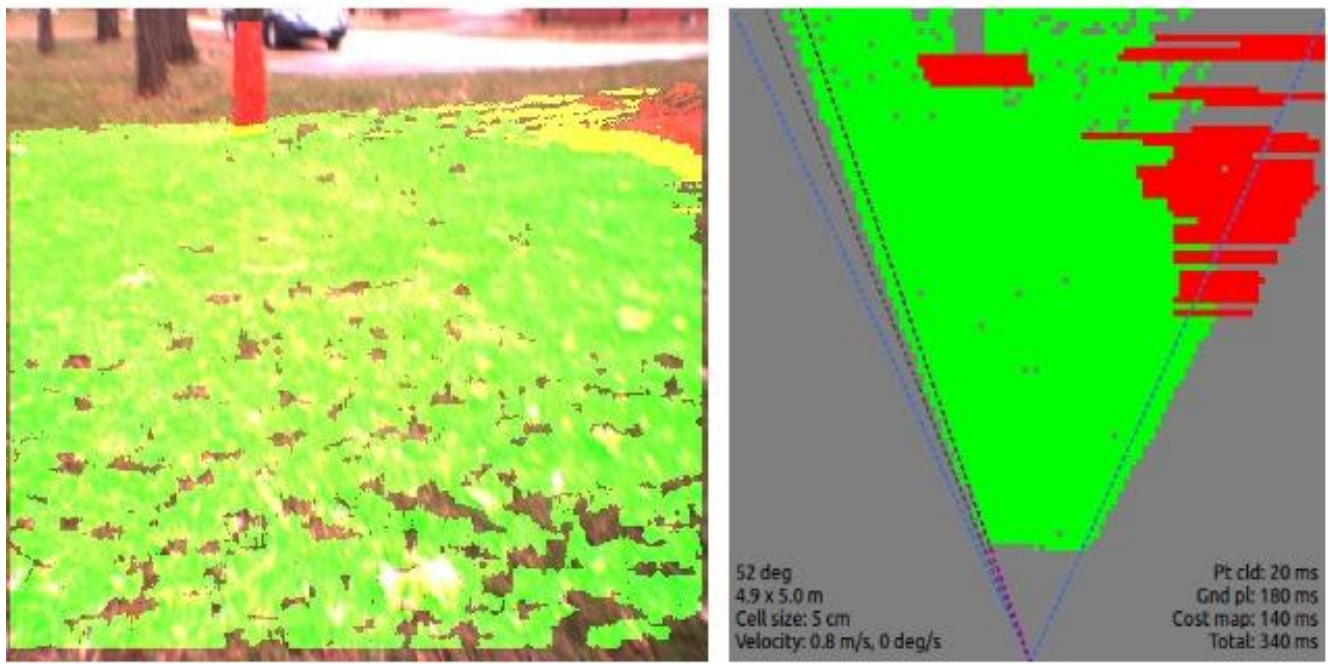

Figure 2. Example of stereo data (left) and corresponding stereo-only cost maps (right). Red indicates detected obstacle and green indicates drivable terrain.

Cost maps obtained solely from 3-D data are often sub-optimal for navigation. For example, knowing the nature of terrain beyond the range of a 3-D sensor is often critical for effective path planning. Furthermore, 3-D data may erroneously label grass as an obstacle and mud as drivable. For these reasons we created a mechanism that allowed us to modify cost maps for particular environments. We used the following procedure:

Images from the camera were scanned creating patches of 59 x 59 pixels. A specialized ConvNet (to be described later) classified the center of each patch as either "drivable" or "obstacle." The classifications thus obtained were used to modify the cost maps - If an obstacle in the cost map was greater than a predetermined threshold height, typically 4 cm, no change was made – we assumed that any obstacle that high was too high for the CoroBot to traverse. For regions in the cost map that corresponded to terrain or obstacles less than threshold height, the cost was determined by the ConvNet. In this way, grass that was under the threshold height could be labeled "drivable", while vehicle-stopping mud puddles (zero height) could be labeled as "obstacle.

# 3 Training the ConvNets

In order to build accurate classifiers, ConvNets with many weights are required. Training these weights requires many labeled training examples. Obtaining an adequate number of training examples using just imagery for the CoroBot camera was not practical since that would have required extensive hand-labelling. Fortunately, the ImageNet database provided an attractive and effective alternative. ImageNet includes about 1,000,000 images that are divided into 1000 labeled classes. In the last few years there has been tremendous progress in creating ConvNets that keep advancing the accuracy of classifying the ImageNet test data. We have been able to piggyback on this progress.

Our systems start with ConvNets that have been trained on ImageNet. We then preserve some of the initial layers of these trained ConvNets for use as feature extractors for our navigation task. A separate navigation training set is then created using the features extracted from the 59 x 59 patches taken from images in our driving environment. We used patches, instead of just single labeled pixels because we hypothesized that having "context" around the pixel in question would improve classification.

A subset of these patches were labeled by a human that identifying areas that might be misclassified by stereo. Next, a single-layer perceptron was trained using these feature/label pairs, i.e. each training example pair contains a feature vector obtained from the feature extractor applied to a patch and a label for that patch which was either "drivable" or "not-drivable". After this training, which can be done in a few seconds, a new classifier was obtained that combined ImageNet-derived feature extraction and customized terrain classification. This new classifier was then used to label terrain in new images without using stereo.

The above strategy is an example of "transfer learning" in which a system designed for one task (in the case ImageNet classification) is repurposed to perform a different task (terrain classification).

## 3.1 ImageNet Learning

The architecture of our ConvNet is shown in Figure 3. During training, the input to our ConvNet has 3 input image RGB planes, each 119 x 119 pixels taken from the ImageNet database.

- The first layer applies 64 filters, each 8x8 pixels, to the input. The first layer used a stride of 4x4, and the maps produced by it are therefore 28x28. This convolutional step is then followed by a thresholding operation, and a max pooling function, which pools regions of size 2x2, and uses a stride of 2x2. The result of that operation is a 64x14x14 array, which represents a 14x14 map of 64-dimensional feature vectors. The receptive field in the input image of each unit at this stage is 8x8.
- The second layer (conv+pooling) is very much analogous to the first, except that now the 64-dim feature maps are projected into 96-dim maps (5x5 size filters). The result of the complete layer (conv+pooling) is a 96x5x5 array.
- In the last stage, the 5x5 array of 96-dimensional feature vectors is flattened into a 2400-dimensional vector, which is fed to a 2-layer classifier with 1536 hidden units, and 1000 output classes.

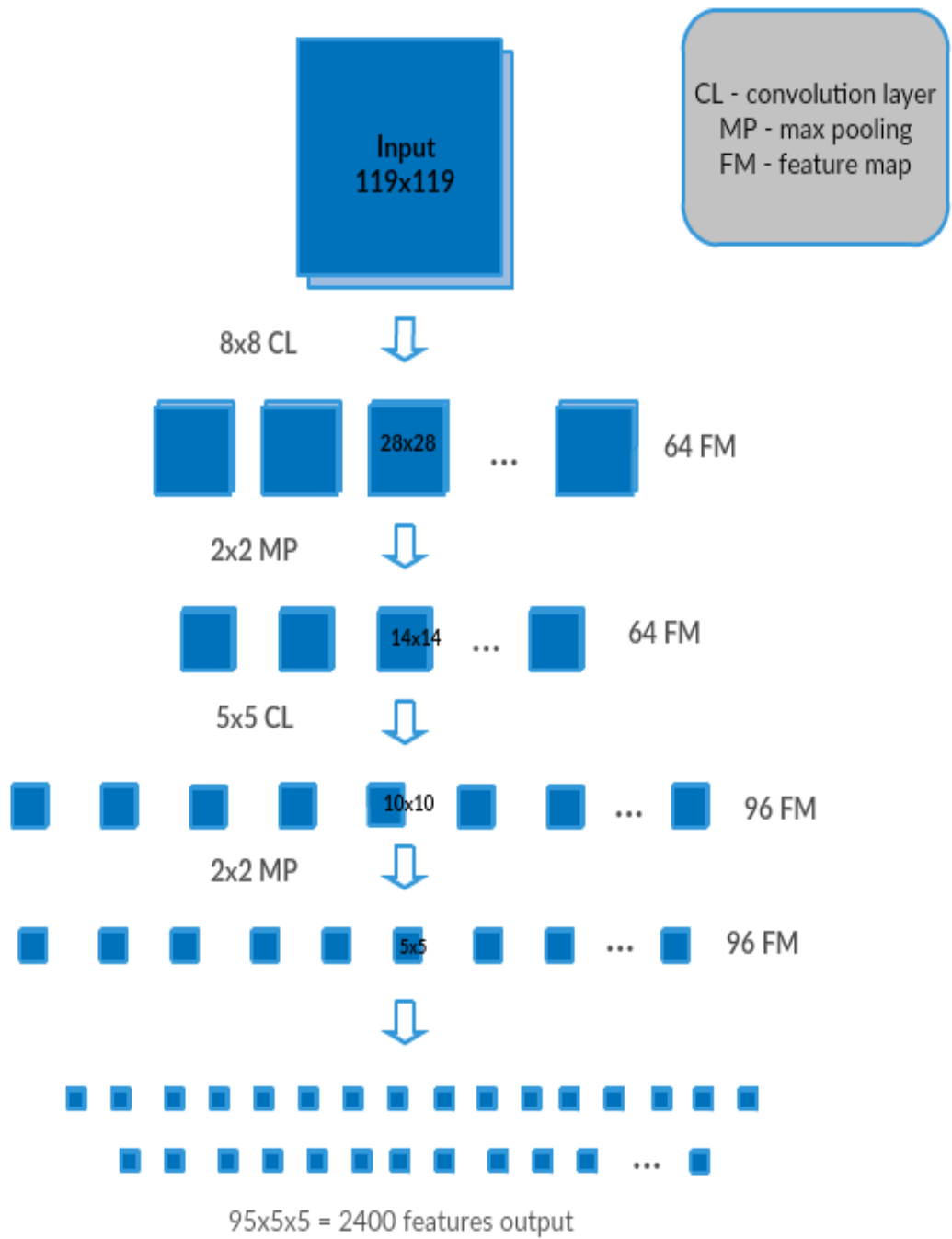

Figure 3. ConvNet feature extractor structure

The loss function is set to negative log likelihood criterion.

### 3.1.1 Details of Training Process

To speed network training, an NVIDIA GTX Titan GPU graphics card was used. To parallelize the training, ~15,000 batches of 64 119 x 119 RGB scaled ImageNet images were prepared as training input along with their ImageNet classification labels. The only pre-processing was subtracting the mean RGB value, computed on the first (shuffled) 10000 images, from each pixel.

We train the network using stochastic gradient descent with momentum set to 0.08. The learning rate is calculated according to the formula $0.05 * \sqrt{batch\ size}/\sqrt{128}$ and multiplied by 0.96 each iteration, which consist of 4 datasets described in the next section. On a system equipped with an NVIDIA GTX Titan GPU, training a single net took one week.

### 3.1.2 Data Set Augmentation

Augmenting the data set by repeatedly transforming the training examples is a commonly employed trick to reduce the network's generalization error. The transformations are applied at training time and do not affect the runtime performance.

We scaled the ImageNet images to 128 x 128 pixels from which we extracted training patches size 119 x 119. Then we randomly rotated, scaled and sheared the training patches according to the following rules:

- Flip: horizontally flip 25% of the examples, vertically flip 25%, both flip 25%, no change 25%.
- Scale: choose a random scale factor in the range of 0.83 to 1.2.
- Rotate: choose a random rotation angle in the range of -30 to +30 degrees.
- Shift horizontal and vertical direction (-5 to 5 pixels) and crop to 119 x 119.

Thus, the training data comprised 4 subsets, each obtained from the ImageNet initial data.

### 3.2 Terrain Classification Task

Our objective here was not to create the very best ImageNet classifier, but rather to learn feature vectors (the output of the penultimate layer in our network) which could then be used for fast training of a linear network that uses navigation sensor input coupled with labeled terrain classes as training examples. With the feature extraction layer weights frozen, and the fully-connected neural network with one or more hidden layers trained using these feature/label pairs appended, we created a classifier to be used during navigation.

Our network was designed to train on regions of an image where the terrain type was known, say from stereo data, or by labeling by a human operator. The networks would then classify regions that were not known, usually from an entirely new frame, but from the same environment.

Images for navigation training were labeled by having a human operator label representative obstacles in the environment. The patches, which were 59 x 59 pixels, were labeled by the class of the center pixel, i.e. each training example contains a feature vector obtained from the feature extractor applied to a patch and the label for that patch "drivable" or "obstacle".

## 4 Implementation Details

### 4.1 Sensors

To process the camera images, we added the Robot Operating System (ROS) image pipeline stack in order to integrate better with the open source ROS environment and to eliminate the dependency of the external and proprietary stereo processing library. This ROS stack provided a set of modules for stereo processing, visualization and camera calibration. The stereo image processing took the raw images from our custom written camera driver for Bumblebee cameras, and performed image rectification (distortion correction and alignment) and color processing for both the left and the right camera. The ROS image pipeline stack performed the required process on the both raw images and finally created the point cloud.

### 4.2 Classification

Torch7 (a scientific computing framework with wide support for machine learning algorithms) was used in the current project because it provided an easy and modular way to build and train simple or complex neural networks. As a part of the navigation system, the classification algorithm was integrated into the ROS environment via a special library that converted images to the Torch tensor representation, facilitating processing by the ConvNet. The ConvNet returned classified labels for each pixel.

The classifier could be trained either in real time or after different images were gathered and labelled by a human. Classifier training could be done on an external computer, i.e. the laptop that is used to control the robot in the field. Training could be accomplished in a few seconds using a GPU enabled laptop. After training, the new classifier was uploaded on to the robot.

## 5 Results and Discussion

### 5.1 Off-line Learning From a Human Teacher

A human teacher was used to label terrain that might be misclassified if we relied solely on stereo. The human labeled representative terrain in one or a few images and we then used those labels for many images in the same environment. Several images were chosen from different scenes as shown in Figure 4**.** Green and Red tints indicate human labeled obstacles and drivable terrain (on the left) and the corresponding classification result (on the right). In the upper images, flat snow-covered ground is classified as an obstacle, while pavement is classified as drivable.

In the lower images of Figure 4 regions with small plants were classified as "drivable" while regions with bushes and trees were classified as "obstacles". While the results are not perfect, they clearly show as "drivable" the terrain that is interspersed with small plants, regions that using stereo alone would have classified as having obstacles.

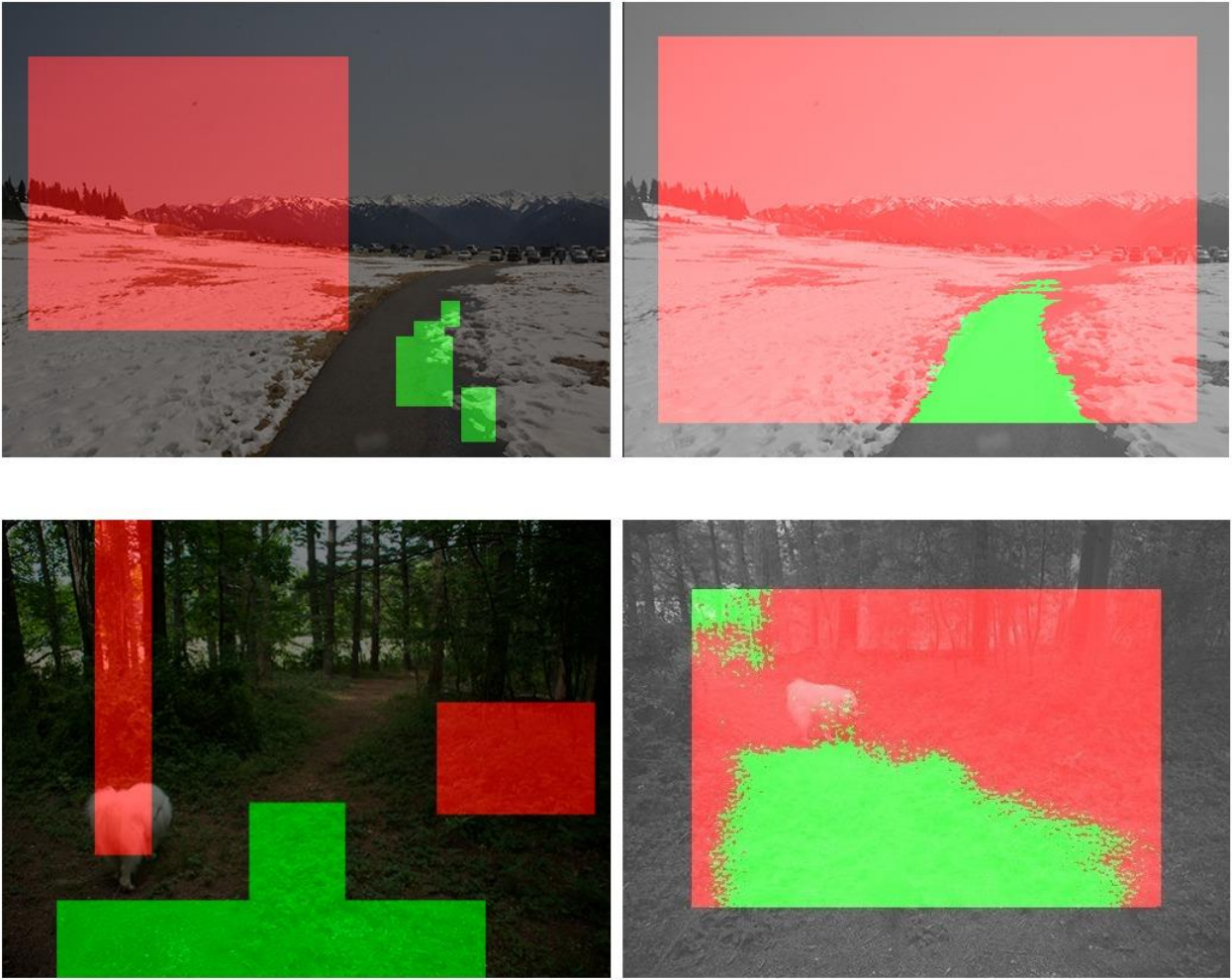

Figure 4: Examples of learning from hand-labeled region (left column) and classifications provided by our system (right column). Red indicates obstacles and green indicates drivable terrain.

Figure 5 shows the effect of adding more human-supplied training labels. The left image in the upper row shows the few pixels that were hand labeled as red and green streaks. The right image shows the classification of the entire image based on these sparse labels. In the lower row, more pixels were labeled (left image) resulting in a more precise labeling of the remaining pixels (right in image).

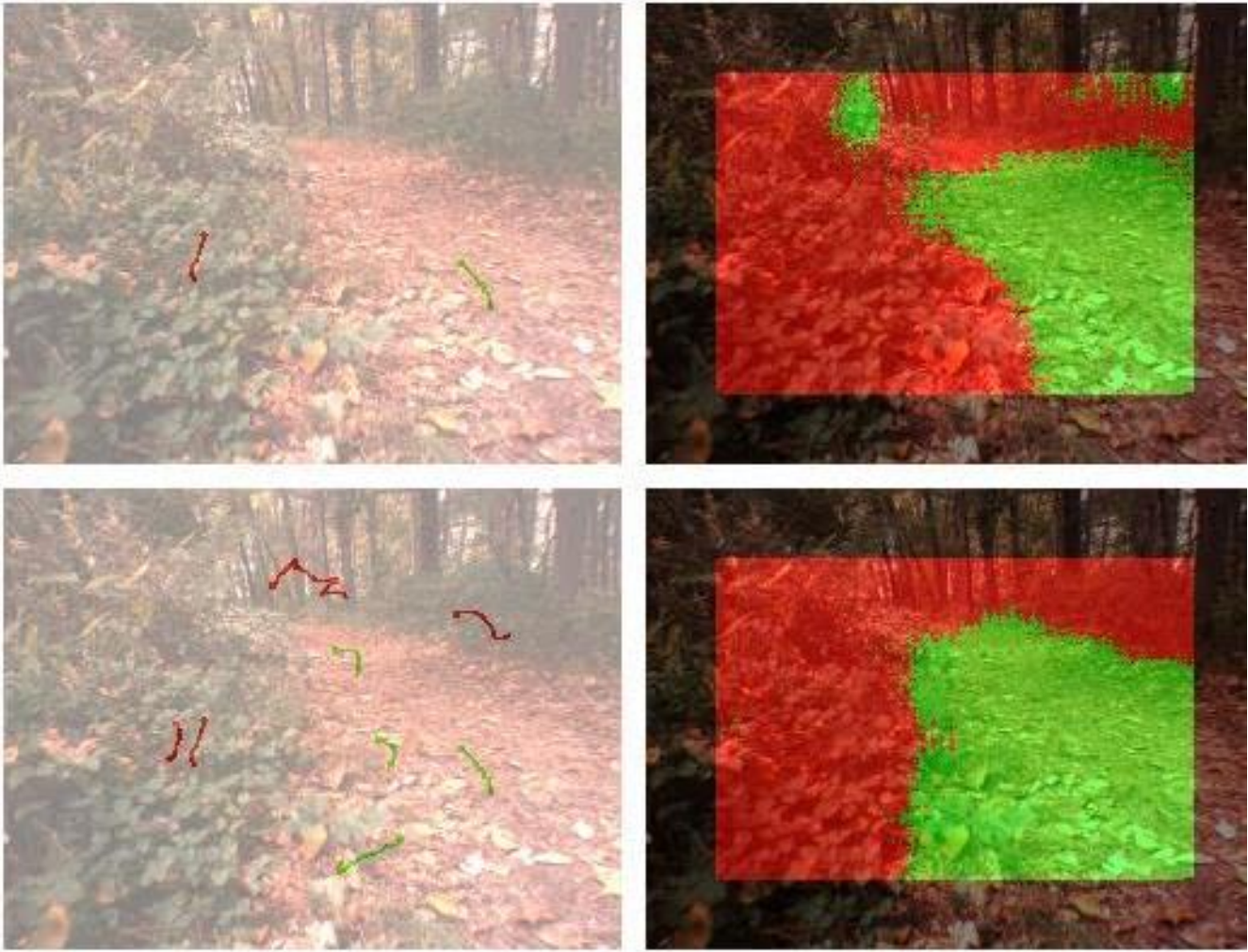

Figure 5. The effect of adding more human-supplied training labels. Labeled areas are shown as overlays in the left column and the resulting classifications are shown in the right column. Red indicates obstacles and green indicates drivable terrain. Adding more labels (lower row), gives a more precise classification than with fewer labels (upper row).

### 5.2 Navigation

The ConvNet could be trained in real time on images from the current environment in which a small number of representative images patches were hand labeled. As noted previously, the patches chosen for hand labeling were ones from regions that would be improperly classified by stereo alone. Labels from the ConvNet and the point cloud were then combined in a cost map using the following method:

First, stereo was used to identify as obstacles objects that extended above the ground plane by more than an operator set threshold, typically 15 cm. For objects less than this threshold, the label obtained from the ConvNet was used for classification. In cases where stereo could not provide a height estimate, such as for surfaces that lack sufficient features to measure disparity, we simply used the labels obtained from the ConvNet.

Below we present results obtained by running the robot in various environments. Corresponding videos can be found on line at

https://www.youtube.com/watch?v=zesKN_1i9VA
https://www.youtube.com/watch?v=AXFupI_Nz8E
https://www.youtube.com/watch?v=gDFQhFbX3oU

### 5.2.1 NYU hall

In this experiment we taught the robot to drive over a carpet while avoiding the rest of the stone floor. The human labeled image is shown in Figure 6.

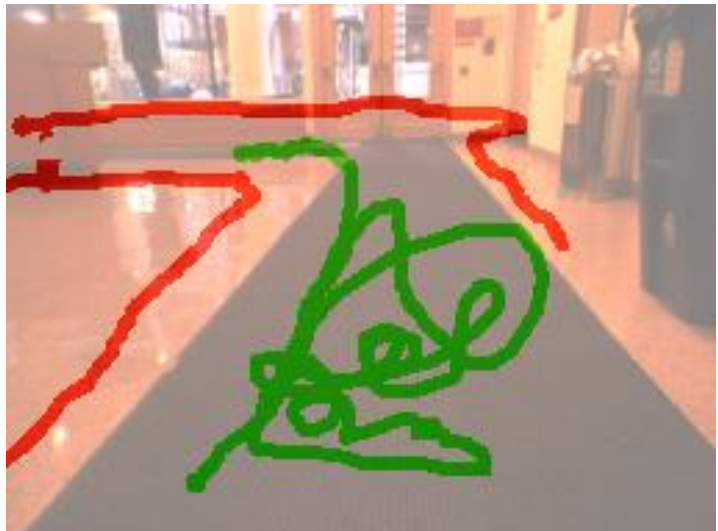

Figure 6. NYU hall with human generated labels for the carpet (green) and stone (red).

As discussed before, once the classifier was trained, a cost map was built from both the learned classifier and from stereo. See Figure 7.

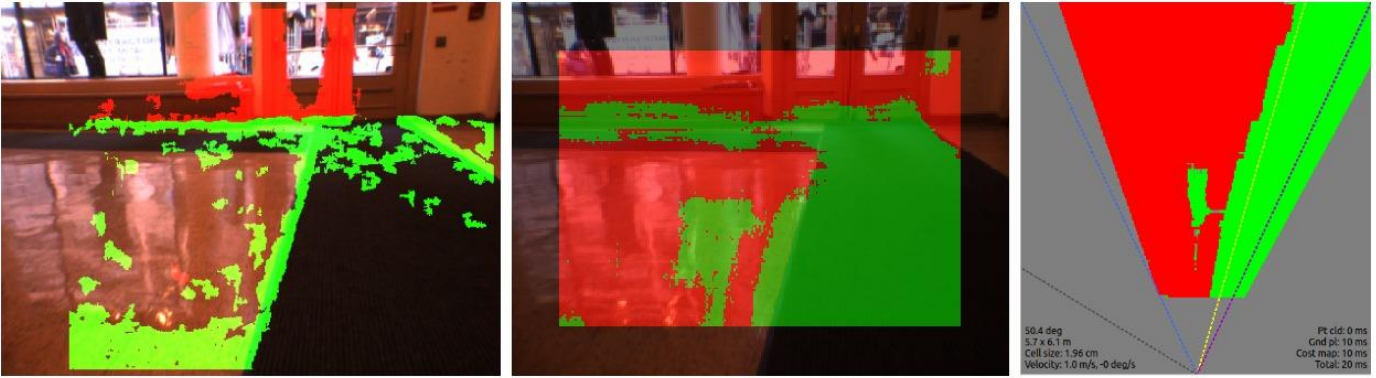

Figure 7: NYU Hall Carpet following from learning

Figure 7 shows classification using only stereo data (on the left), the classified image using the human supplied labels from Figure 6 (middle) and the corresponding cost map (on the right).  In the cost map, the red areas (obstacles) were widened by 15 cm in all directions to account for the width of the robot.  Though classification was sensitive to shadows and lighting conditions, the cost map was much more complete than one that would be generated by stereo alone, since stereo data is not reliable on surfaces without significant texture, such as the highly polished floor tiles. In addition, the map reflects the labeling choices desired by the human.

### 5.2.2 Bayonet Farm

In this experiment we show an outdoor scene (Figure 8) with human labeling, using a GPS sensor for localization.

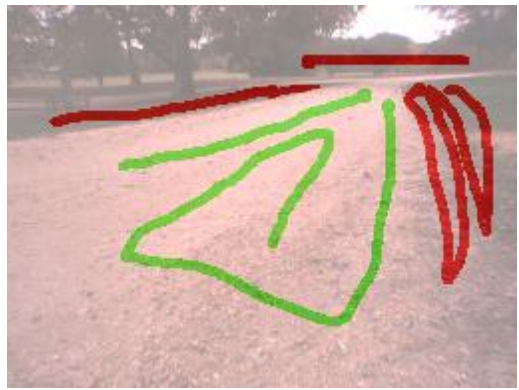

Figure 8. Bayonet farm labels of path (drivable) and grass and vegetation (obstacle)

Here stereo gave us reliable information, but we didn't want to drive over the grass. By training a classifier and learning human labels, we built the desired cost map for navigation. See Figure 9. The left image in the figure shows terrain labeling from stereo in which most of the ground is labeled drivable. The middle image shows terrain labeling using the human supplied labels of Figure 9.  The right image shows the resultant cost map.

Page 11 of 14

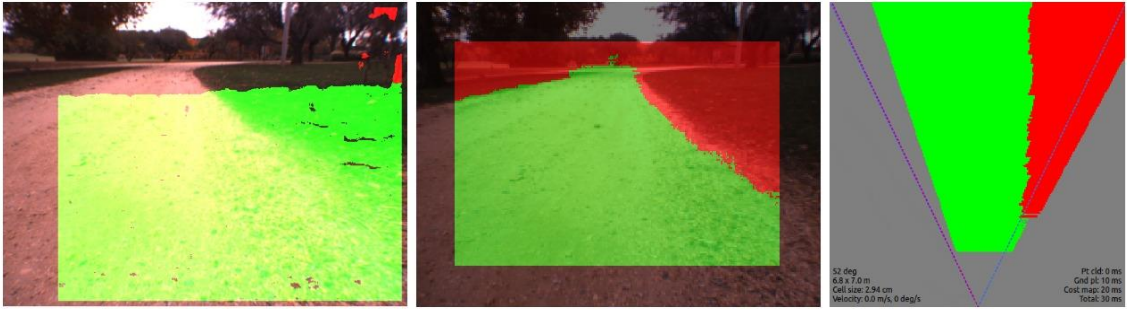

Figure 9. Bayonet Farm path following from learning.

**5.2.3 Net-Scale Backyard**

In this experiment we illustrate a case where stereo identifies grass as an obstacle but we want the robot to drive on grass as well as on the cement walkway. Figure 10 shows the human supplied labels for this scene.

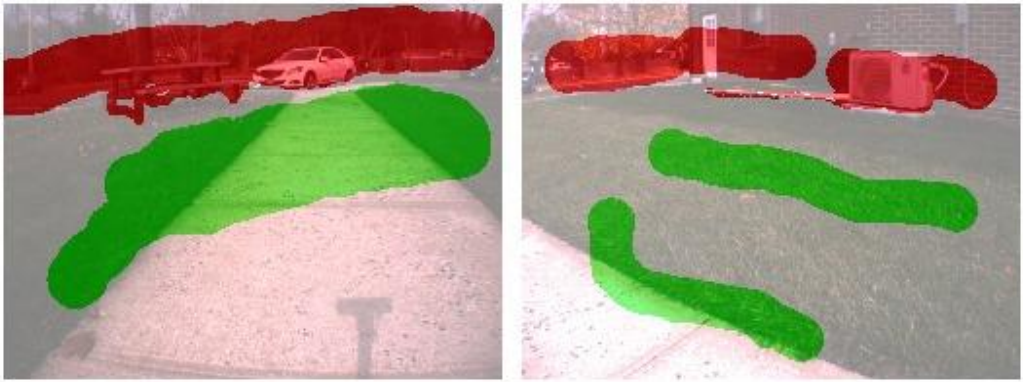

Figure 10. Net-Scale backyard human labels for the path

Figure 11 shows labeled images and corresponding cost maps. The upper row shows labeling based only on stereo; the lower row shows labeling based on stereo and human labels. Comparing the two cost maps, we see that after learning the human supplied labels, the map has its desired appearance, with both the cement and grass labeled as drivable.

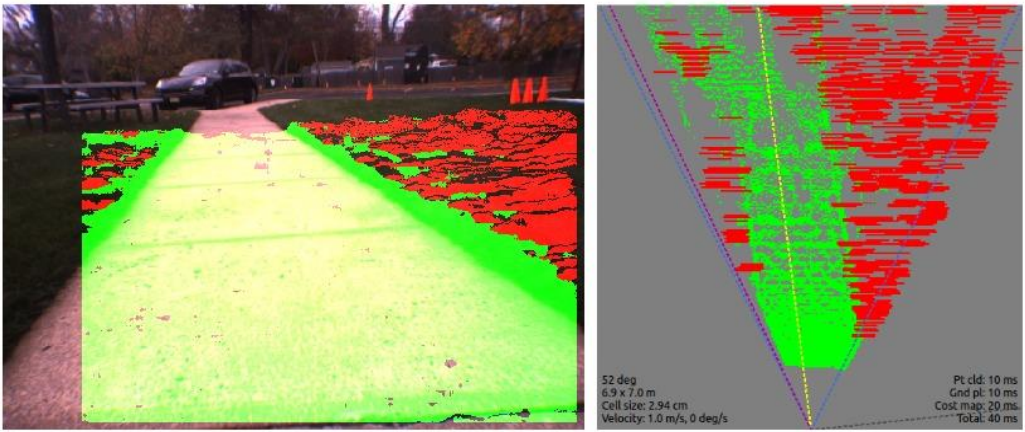

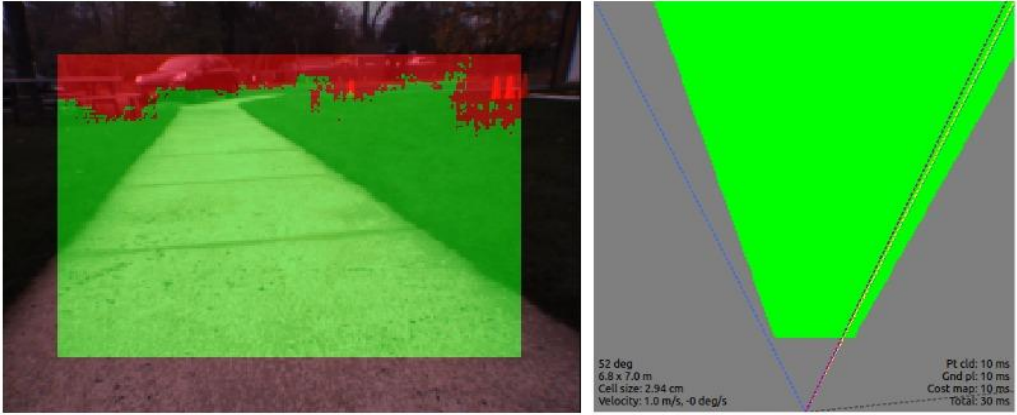

Figure 11: Net-Scale backyard path.

## 6 Conclusion

We have shown that using the feature extractors obtained from previously trained ConvNets provides good terrain classification, even though the original ConvNets were trained on a completely different corpus. Training the single layer perceptrons that use the feature vectors extracted by the initial layers of the ConvNets is sufficiently fast that the process provides a promising method for rapid adaptation while navigating in diverse environments.

These results illustrate the feasibility of real-time tuning of autonomous robot navigation preferences, creating a system that is highly customizable and adaptive.

**Acknowledgement**
This material is based upon work supported by the United States Army under Contract No. W56HZV-13-C-0014.

Any opinions, findings and conclusions or recommendations expressed in this material are those of the author(s) and do not necessarily reflect the views of the United States Army